\documentclass[journal]{IEEEtran}

\usepackage{graphicx}
\usepackage[inline]{enumitem}
\usepackage{cite}
\usepackage[dvipsnames]{xcolor}
\usepackage{xr}
\usepackage{amsmath}
\usepackage{subcaption}
\usepackage{siunitx}
\usepackage{multirow}
\usepackage{slashbox}

\usepackage[bookmarks=true]{hyperref}
\usepackage{bm}

\DeclareMathOperator*{\minimize}{Minimize}
\DeclareMathOperator*{\subjectto}{Subject\ to}
\newcommand{\norm}[1]{\left\lVert#1\right\rVert}

\begin{document}

\title{Optimization-Based Framework for Excavation Trajectory Generation}
%
%
%

\author{Yajue Yang$^{1}$, Pinxin Long$^{2}$, Jia Pan$^{3}$, Xibin Song$^{2}$,  Liangjun Zhang$^{2}$
\thanks{$^{1}$Yajue Yang is with the Department of Mechanical and Biomedical   Engineering, City University of Hong Kong, Hong Kong, China}
\thanks{$^{2}$Pinxin Long, Xibin Song and Liangjun Zhang are with Baidu Research, Baidu Inc., Beijing, China}
\thanks{$^{3}$Jia Pan is with the Department of Computer Science, the University of Hong Kong, Hong Kong, China}
}

%


\maketitle

\begin{abstract}

In this paper, we present a novel optimization-based framework for autonomous excavator trajectory generation under various objectives, including minimum joint displacement and minimum time. Traditional methods on excavation trajectory generation usually separate the excavation motion into a sequence of fixed phases, resulting in limited trajectory searching space. Our framework explores the space of all possible excavation trajectories represented with waypoints interpolated by a polynomial spline, thereby enabling optimization over a larger searching space. We formulate a generic task specification for excavation by constraining the instantaneous motion of the bucket and further add a target-oriented constraint, i.e. swept volume that indicates the estimated amount of excavated materials. To formulate time related objectives and constraints, we introduce time intervals between waypoints as variables into the optimization framework. We implement the proposed framework and evaluate its performance on a UR5 robotic arm. The experimental results demonstrate that the generated trajectories are able to excavate sufficient mass of soil for different terrain shapes and have $60\%$ shorter minimal length than traditional excavation methods. We further compare our one-stage time optimal trajectory generation with the two-stage method. The result shows that trajectories generated by our one-stage method cost $18\%$ less time on average. 

\end{abstract}


%
\IEEEpeerreviewmaketitle

\section{Introduction}
\label{sec:intro}
Heavy machinery, such as hydraulic excavator, is widely used in construction, mining and many other scenarios. Nowadays, excavators are mainly controlled 
by humans operators. In order to address the labor shortage for skillfull operators, save the increasing labor cost, and improve working conditions especially for hazardous environment, strong need is existed to develop autonomous excavators and automate the overall excavation process. Recently, albeit several prototyping systems \cite{singh1992task,sing1995synthesis,stentz1999robotic, ha2002robotic,schmidt2010simulation,jud2017planning,jud2019autonomous} have been developed to automate excavation motions, however, rarely any autonomous excavators have been deployed in real-world service.

One major challenge for developing autonomous excavator is to generate feasible and optimal trajectories for excavator to execute. An excavation task is composed by repetitive motion cycles, and a single cycle of excavation is defined as generating a feasible motion trajectory and executing it to excavate the given volume of soil without violating the specified constraints. 
To maximize the productivity and minimize the fuel consumption, it is desired to find an optimal trajectory to perform the excavation cycle under the predefined objectives and constraints. Furthermore, excavation tasks require special attentions for time-related objectives and constraints because excavators are required to synchronize with other machines in many scenarios for efficient collaboration, such as earth movers and loading trucks.



\begin{figure*}
\captionsetup[subfigure]{position=b}
\centering
\subcaptionbox{bucket shape\label{fig: bkt_shape}}{\includegraphics[width=.239\textwidth]{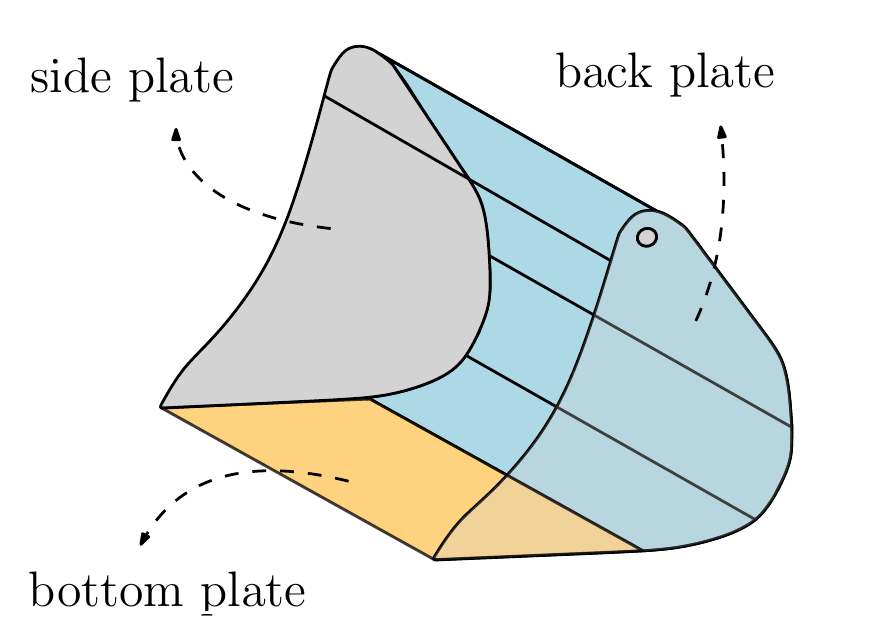}}
\subcaptionbox{penetration mode\label{fig: pen_mode}}{\includegraphics[width=.239\textwidth]{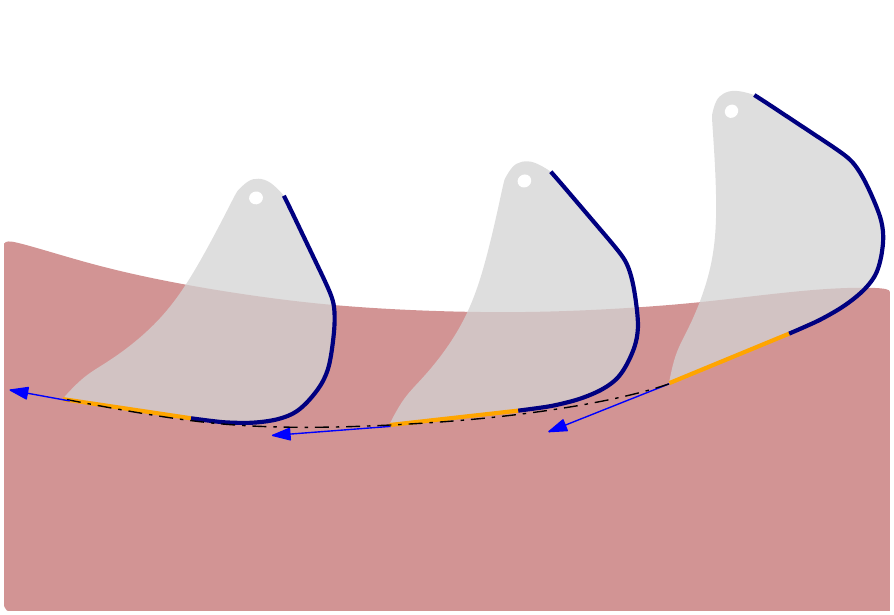}}
\subcaptionbox{separation mode\label{fig: sep_mode}}{\includegraphics[width=.239\textwidth]{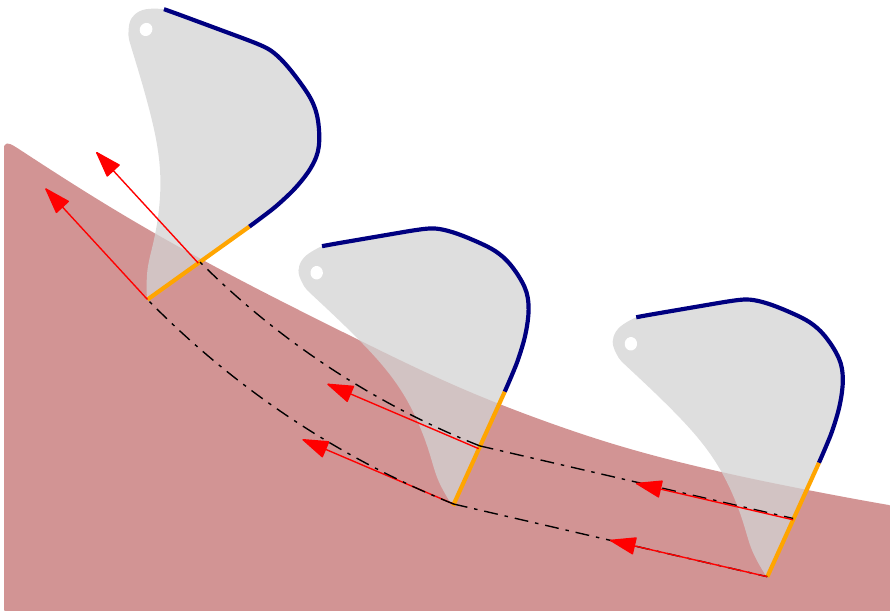}}
\subcaptionbox{composite mode\label{fig: cps_mode}}{\includegraphics[width=.239\textwidth]{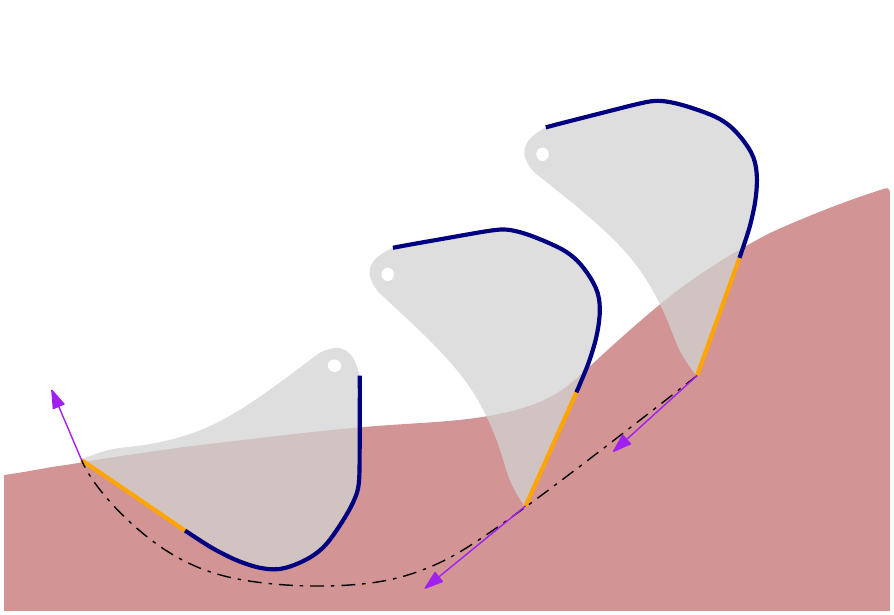}}
\caption{(a) Typical bucket shape. From (b) to (d), we depict different trajectories capable of excavation. (b) Along the trajectory, the bucket penetrates through the soil along the direction tangent to the bottom plate. The tangential velocity is denoted with the blue arrow. (c) Along the trajectory, the bucket breaks and moves the soils in front of the bottom plate by pure separation~\emph{i.e.} moving in normal direction of the bottom plate. The normal velocity is denoted with the red arrow. (d) Along this trajectory, bucket velocity at each timestep is composed of the tangent and normal velocity. The resultant velocity is denoted with the purple arrow.
}
\label{fig: digging_modes}
\end{figure*}

Traditional methods on trajectory generation for excavation usually separate the bucket path into three phases: linear penetration, dragging and scooping, of which the first two are restricted to pure translation movement for simplicity~\cite{sing1995synthesis,sotiropoulos2019model,sandzimier2020data}.
However, subgroups of possible excavation trajectories are concerned in these methods, which heavily reduces the space for optimization. 
As illustrated in Fig.~\ref{fig: digging_modes}, essentially, the key point of digging soils out is that the motion of the robot bucket can cause the soil failure phenomenon (i.e. a mass of soil failing to retain its original geometric shape and displacement~\cite{mckyes1985soil}).
As depicted in Fig.~\ref{fig: pen_mode} and Fig.~\ref{fig: sep_mode}, two primitive \textit{instantaneous} motions of a typical bucket are existed to cause the failure of soils in front of the bottom plate:
\begin{enumerate*}
    \item the bottom plate edge (or teeth) penetrates into the soil along the direction tangent to it, which leads to a rupture of soils contacting with the bottom plate;
    \item the bottom plate separates and presses the soil material in front of it with the normal forces, which leads to a rupture of a wedge area ahead of it~\cite{park2002development}.
\end{enumerate*}
Digging trajectories are varied because the bucket motion at each timestep during the entire excavation process could combine the instantaneous penetration and separation in many different ways. Fig.~\ref{fig: cps_mode} gives an example of a valid excavation trajectory with different composite motions at each timestep. While previous methods stick to the rigid~\emph{three-phase} pattern for excavation trajectory, we propose to regulate the trajectory in a more flexible way by constraining \emph{instantaneous} motions and their relationships, which enables optimization over a wider trajectory space.

In this paper, we formulate the excavation task as a trajectory optimization problem with end-effector (\emph{i.e.} bucket) constraints, on which a great deal of research is conducted.
Many sampling-based planning algorithms are developed whose core idea is to incorporate constraints into planning by projecting sampled configuration states onto the implicit manifold defined by an equality constraint function~\cite{kingston2020decoupling}. Although successful for many robotic tasks, they are not suitable for the excavation task because the constraints are too complex to be expressed with an equality function. Another approach is the trajectory optimization, which formulates the task as a general optimization problem over the trajectory space and finds the optimal trajectory under task-specific constraints~\cite{zucker2013chomp}. 
We adopt this approach to handle a set of complex constraints with advanced optimization techniques such as sequential quadratic programming~\cite{schulman2014motion} and stochastic initial trajectory generation~\cite{kalakrishnan2011stomp}.


Considering time related objectives and constraints, we introduce time intervals as variables into the optimization problem, which enables simultaneous generating a constraint satisfied path and minimizing the travelling time. Different from this method, the classical method of time optimal trajectory generation is to separate this process into two steps, which is known as the \textit{two-stage} approach~\cite{bobrow1988optimal}. 
It first produces a geometric path, and then generates the time-optimal trajectory that exactly follows the path under velocity and acceleration limits or constraints induced by dynamics singularities~\cite{kunz2012time, pham2014general,hauser2014fast}. 
For tasks without a predefined path, such as the excavation task, one can try the two-stage method multiple times with various geometric paths and select the optimal one~\cite{pham2014general}. As a comparison, we refer our method as~\emph{one-stage} method in the rest of the text. We adopt the~\emph{one-stage} method since it is potential to find better trajectories with advanced optimization techniques~\cite{rosmann2020time, cao1997constrained}.
\textbf{Main Contributions:} In summary, this paper presents a novel optimization-based framework of excavation trajectory generation. Specifically, a continuous trajectory in joint configuration space is represented with discrete waypoints interpolated by a polynomial spline. Apart from the waypoints, time intervals are introduced as variables in this optimization problem, which facilitates explicitly arranging time profile of trajectories. A group of process-oriented geometric constraints are imposed to ensure that the bucket motion obeys the excavation principle. To collect sufficient amount of soil, we add a target-oriented constraint,~\emph{i.e.} swept volume factor that indicates the estimated amount of excavated materials. This constraint-based task specification guarantees the success of excavation and allows large room for trajectory optimization. The framework is suitable for various objective functions~\emph{e.g.} minimum time, minimum joint displacement and minimum torque, etc.

We implement the proposed framework on a UR5 manipulator arm to demonstrate that the generated trajectories are able to excavate sufficient mass of soil for terrains in different shapes. We first show that the minimal length of trajectories generated by our method are $60\%$ shorter than the aforementioned three-phase methods. Then we compare our one-stage method with the two-stage method in time optimal excavation trajectory generation. The experiment result shows that trajectories generated by the one-stage method cost $18\%$ less time on average. 

The rest of the paper is organized as follows. Section~\ref{sec:related_work} gives an overview of previous work. Section~\ref{sec: method} presents the optimization framework including the problem statement, task-specific constraints, the usage of interpolation spline for path constraints and the optimization formulation. Empirical experiment results are shown and analyzed in Section~\ref{sec: exp}.

\section{Related Work}
\label{sec:related_work}

Many previous studies on excavation are developed for various motivations based on the classical three-phase trajectory characterization.
Observing the fact that the capability of translation is limited by the nature of excavators' mechanism, Yang et al. proposed a compact reachability map of excavators which facilitates searching feasible rectilinear motions for penetrating and dragging~\cite{yang2019compact}.
Jud et al. developed a controller to exert a desired bucket force at every instantaneous step in the operational space~\cite{jud2017planning}, in which desired force profiles are manually designed to comply with the three-phase pattern. This method enables trajectories adaptive to different soils by switching to the next phase once the bucket halts due to large resistance force from the terrain.
Sandzimier et al. proposed a data-driven statistical model of predicting the amount of soil collected with Gaussian processes~\cite{sandzimier2020data, sotiropoulos2020autonomous}. According to the prediction model, the method controls the bucket to switch from the drag phase to the rotation phase at a proper time when the desired amount of soil would be excavated.
In contrast to these local control strategies, our work jumps out of three-phase framework and focuses on global optimization over the entire trajectory.

Kim et al.~\cite{kim2013dynamically} generated time-efficient and minimum torque motions for excavators with parameter optimization method. Since they regard the digging process as a point-to-point motion without explicitly imposing path constraints, the generated trajectory is prone to undesired movement such as pressing soils. Furthermore, the interaction force model they adopt is only suitable for horizontal translation and thus is inaccurate when the model is applied for arbitrary point-to-point motions.
Park et al.~\cite{park2002development} develops an algorithm to select one digging mode out of several candidates and generate the corresponding joint trajectory in a hard-coded way.

\section{Method}
\label{sec: method}
 In this section, we present our time variable optimization-based framework of excavation trajectory generation. We represent the trajectory with a waypoint-interpolating spline. To make the trajectory suitable for the excavator to execute, we impose task specific constraints on the set of waypoint and the intermediate points sampled along the spline. Note that only the waypoints are variables in the optimization problem, which control the shape of the spline and any points on the spline can be interpolated as a function of the waypoints.

\subsection{Problem Statement}
A trajectory $\mathbf{q}(t) : \mathcal{R} \mapsto \mathcal{C} \subseteq \mathcal{R}^d$ is a continuous function mapping time  $t$ to a $d$-dimensional robot configuration state $\mathbf{q}$ within a constrained region $\mathcal{C}$. Trajectory optimization refers to minimizing some objective functionals over a set of constraints over the trajectory:
\begin{align}
    \minimize_{\mathbf{q}(t)}\quad &\int_0^{t_f}L(\mathbf{q}(t))dt && \label{eq: opt_ctrl_cost} \\
    \subjectto\quad &\mathbf{g}(\mathbf{q}(t)) \leq \mathbf{0} &&\\
    &\mathbf{h}(\mathbf{q}(t)) = \mathbf{0}&&
    \label{eq: opt_ctrl_cstr}
\end{align}
where the final time $t_f$ could be either fixed or flexible~\cite{quintana1973numerical}.
To make the problem tractable, $\mathbf{q}(t)$ is discretized into a finite sequence of waypoints: $ \mathbf{Q} = \{\mathbf{q}_0, \mathbf{q}_1, \dots, \mathbf{q}_n\} $ with the time knots ranging from $t_0$ to $t_n$. With $t_0 = 0$, we introduce $n$ time intervals as optimization variables $T = \{\Delta t_0, \Delta t_1, \dots, \Delta t_{n-1}\}$ where $\Delta t_i = t_{i+1} - t_i$. Consequently, the original problem is converted to a nonlinear programming over a multidimensional vector which is comprised of $d \times (n+1)$ joint position variables and $n$ time variables.

\begin{figure}
    \centering
    \includegraphics[width=.75\linewidth]{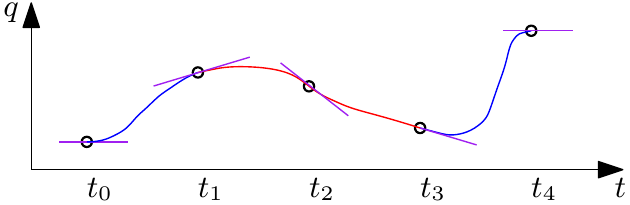}
    \caption{Spline combining quitic polynomials and one exact cubic spline. The red spline denotes an exact cubic spline and blue curves are quintic splines. Purple lines represent the first derivatives at waypoints.}
    \label{fig: qc_spline}
\end{figure}

We use a second-order differentiable polynomial spline $\mathbf{s}(t)$ interpolated between the discrete waypoints to represent an excavation trajectory. Although a cubic spline is sufficient for the requirement of smoothness, we adopt a mixed-order spline: the two segments at the ends are interpolated with quintic polynomials while the remaining waypoints are interpolated with a cubic spline, as shown in Fig.~\ref{fig: qc_spline}. In particular, the spline $\mathbf{s}(t)$ is formulated as follows:
\begin{itemize}
    \item for $i=0$ and $i=n-1$,
\begin{equation}
    \mathbf{s}(t) = \sum_{j=0}^{j=5}\mathbf{a}_{ij}(t - t_i)^j,\quad \text{if } t_i \leq t < t_{i+1}
\end{equation}
\item for $0 < i < n-1$,
\begin{equation}
    \mathbf{s}(t) = \sum_{j=0}^{j=3}\mathbf{a}_{ij}(t - t_i)^j,\quad \text{if } t_i \leq t < t_{i+1}.
\end{equation}
\end{itemize}

\begin{figure}
    \centering
    \includegraphics[width=.65\linewidth]{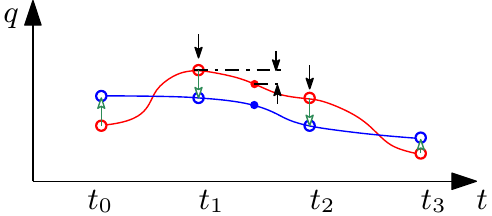}
    \caption{Illustration of path constraints enforcement. Spline curves interpolate all the waypoints marked by hollow circles and the dotted circles are other intermediate points on the spline.
    We use red and blue colors to distinguish the first and second iterations. Constraints on the waypoints and intermediate points are imposed at the first iteration and denoted by black arrows, which push the waypoints at $t_1$ and $t_2$ downwards and reduces the difference in value between a point between them and the waypoint at $t_1$. After performing update to the four waypoints as denoted by the green arrows, we get the blue spline satisfying the constraints. }
    \label{fig: spline_opt}
\end{figure}

Utilizing higher order polynomials helps in making the spline more versatile to satisfy complex constraints while keeping a small number of waypoints, which is crucial for optimization convergence.
The spline parameters  $\mathbf{a}_{ij}$s and any point on the spline can be one-to-one determined by the waypoints $\mathbf{Q}$ and time intervals $T$. Thus, path constraints on arbitrary spline points can be transformed to constraints on the optimization variables $\mathbf{Q}$ and $T$. Fig.~\ref{fig: spline_opt} illustrates how a path constraint is enforced in an optimization process.

The integral objective function in Equation~\ref{eq: opt_ctrl_cstr} can accordingly be transformed into sum of terms. For example, the length of trajectory can be represented as:
\begin{equation}
    f(\mathbf{Q}, T) = \sum_{i = 0}^{m}\norm{\mathbf{s}(t_{i+1}) - \mathbf{s}(t_i)}^2
    \label{eq: jnt_pos_obj}
\end{equation}
where $\mathbf{s}(t_i)$ denotes the point sampled on the spline at $t_i$ and $m$ is the number of samples. In the similar way,
the time-related objective and constraint can also be formulated as:
\begin{equation}
    \minimize_T\ \sum_{i = 0}^{n-1}\Delta t_i \quad \text{and} \quad \sum_{i = 0}^{n-1}\Delta t_i = t_f
\end{equation}
where $t_f$ denotes the desired finish time.




\subsection{Excavation Task Specific Constraints}
\begin{figure}
    \centering
    \includegraphics[width=.7\linewidth]{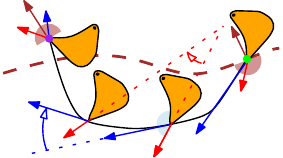}
    \caption{Geometric constraints of the underground trajectory for valid excavation. The black curve denotes the bucket path, starting from the green point to the purple point which are anchor points. Blue arrows, tangential directions of the curve, denote translation directions of the bucket tip. Red arrows represent heading directions of the bucket. The normal vector of terrain surface, depicted with the brown arrow, confines heading and translation directions at both anchor points in the area denoted with the brown region. As the hollow arrows describe, heading and translation directions are constrained to monotonously rotate clockwise along the path. At every point, the hyperplane separated by the heading direction where the translation direction is limited to lie on is denoted with the light blue region.
    }
    \label{fig: excvt_cstr}
\end{figure}

The principle of a successful excavation is that the bucket continuously breaks the soils in front of the bottom plate and lifts them out of the terrain when the broken soils accumulate enough. As mentioned in Section~\ref{sec:intro}, instantaneous motions that contribute to causing failure of front soils are classified into three categories: penetration, separation and the composite motion. Essentially, a digging operation is a sequential combination of such instantaneous motions with suitable geometric guidance and regulation to achieve the goal of excavation. Based on this fact, below we propose a list of rules of thumb on the excavation process, which constrains the instantaneous motions. For convenience, we refer to the direction of the bottom plate as heading direction in the rest of this paper. As illustrated the Fig.~\ref{fig: excvt_cstr}, \emph{Rules} for excavation include:
\begin{enumerate}
    \item At both ends of the trajectory, the bucket tip position is constrained to attach the terrain surface;
    \label{rule1}
    \item At the first point, the bucket tip translation direction (as described in Fig.~\ref{fig: excvt_cstr}) is constrained to be opposite to the normal vector of the terrain surface to enter into the terrain;
    \label{rule2}
    \item At the last point, translation direction should be along the normal vector of the terrain to leave;
    \label{rule3}
    \item From the beginning to the end, the translation direction rotates monotonically;
    \label{rule4}
    \item From the beginning to the end,  the translation direction points to the left half side of the heading direction.
    \label{rule5}
    \item At the first point, the heading direction is constrained to be opposite to the normal vector of the terrain surface for breaking into the terrain surface;
    \label{rule6}
    \item At the last point, the heading direction is constrained to be upward so that the collected soils do not spill out;
    \label{rule7}
    \item From the beginning to the end, the heading direction must either stay the same or rotate clockwise, otherwise the bucket back will press the soil;
    \label{rule8}
\end{enumerate}
\emph{Rule}~\ref{rule1} to \emph{Rule}~\ref{rule4} serve to constrain the bucket path to be under the terrain surface and in a roughly concave shape. \emph{Rule}~\ref{rule5} serves to regulate instantaneous motions beneath the terrain to cause soil failure. \emph{Rule}~\ref{rule6} to \emph{Rule}~\ref{rule8} together serve to constrain the bucket to rotate gradually to sweep soils up in it. 
Note that \emph{Rules} are sufficiently relaxed in the sense that they do not rigidly restrict the motion pattern as previous work does, thereby allowing more diverse trajectories and leaving a larger room for optimization.

In summary, \emph{Rules} mainly stipulate how the bucket move to break and collect soils by imposing constraints on bucket poses and motions along the path. However, they do not specify the amount of soils to be excavated.  Therefore, we introduce another constraint, called swept volume constraint, which estimates the soils to be excavated with the volume of soils above the bucket path. Apart from these excavation specific constraints, common robotic constraints such as joint velocity and acceleration limits are imposed in this problem. In the following text, we first explain constraints on the pose and motion of the bucket according to~\emph{Rules}. Then we describe the swept volume constraint.

\subsubsection{Constraints on Bucket Pose and Motion}

A bucket pose is obtained through the forward kinematics (FK) mapping:
\begin{equation}
    \text{Pose} = (\bm{p}, \Vec{\bm{h}}) = \textsc{FK}(\bm{q})
\end{equation}
where $\bm{p}$ denotes the position of the bucket tip, $\Vec{\bm{h}}$ denotes the heading direction. We add an arrow symbol on a direction variable to distinguish a free vector from a multi-dimensional point. Bucket translation $\Vec{\bm{t}}$ between two poses is then obtained by subtraction between tip positions. By $\Vec{\bm{t}}$ we want to emphasize that what matters is the direction although a translation vector indeed has length. As \emph{Rules} show, we do not impose any constraint on the length of translation.

All constraints on directions in this work are formulated in terms of the hyperplane. A hyperplane passing through the origin is defined as a linear equation
\begin{equation}
    \text{hyperplane} = \{\bm{x}\ |\ \bm{x} \cdot \Vec{\bm{n}} = 0\}
\end{equation}
where $\Vec{\bm{n}}$ denotes the normal vector.
A hyperplane divides the space into two half-planes, one of which is defined as
\begin{equation}
    \text{halfplane} = \{\bm{x}\ |\ \bm{x} \cdot \Vec{\bm{n}} < 0\}
\end{equation}
The region that a direction is confined to can be a half-plane or a intersection of half-planes. Take the case in Fig.~\ref{eq: dir_cstr} as an example, the region is formulated as
\begin{equation}
    \begin{cases} 
          &\Vec{\bm{v}} \cdot \Vec{\bm{n}}_1 < 0 \\
          &\Vec{\bm{v}} \cdot \Vec{\bm{n}}_2 < 0
    \end{cases}
    \label{eq: dir_cstr} 
\end{equation}
It is clear that \emph{Rule}~\ref{rule2},~\ref{rule3},~\ref{rule6} and~\ref{rule7} can be formulated in a similar way to Eq.~\ref{eq: dir_cstr} with $\Vec{\bm{n}}$ determined by the terrain surface.

As to path constraints prescribing bucket motions and their relationship, we add constraints to $m$ points uniformly sampled on the path. Constraints involving monotonous direction changes can be formulated as:
\begin{equation}
    \text{for } 0 < i < m-1,\quad \Vec{\bm{v}}_{i+1} \cdot \Vec{\bm{n}}_{\Vec{\bm{v}}_i} < 0
\end{equation}
where $\Vec{\bm{n}}_{\Vec{\bm{v}}_i}$ denotes the normal vector corresponding to $\Vec{\bm{v}}_i$. $\Vec{\bm{v}}$ can be replaced by $\Vec{\bm{t}}$ for~\emph{Rule}~\ref{rule4} and $\Vec{\bm{h}}$ for~\emph{Rule}~\ref{rule8}. Similarly, \emph{Rule}~\ref{rule5} concerning the relationship between translation direction and heading direction is formulated as:
\begin{equation}
    \text{for } 0 < i < m-1,\quad \Vec{\bm{t}}_{i} \cdot \Vec{\bm{n}}_{\Vec{\bm{h}}_i} < 0
\end{equation}

\subsubsection{Swept Volume Constraint}
\begin{figure}
\captionsetup[subfigure]{position=b}
\centering
\subcaptionbox{\label{fig: dir_cstr}}{\includegraphics[width=.49\linewidth]{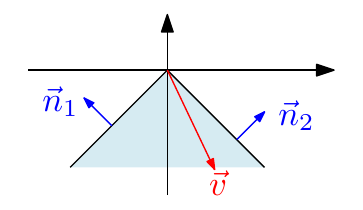}}
\subcaptionbox{\label{fig:sv_illustration}}{\includegraphics[width=.49\linewidth]{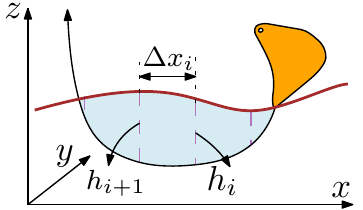}}
\caption{Calculation of constraints. (a) Illustration of the direction constraint. $\Vec{\bm{v}}$ is restricted to be located between the dash lines. $\Vec{\bm{n}}$s denote normal vectors. (b) Swept volume estimation. Brown curves denote terrain surfaces. Black curves are the bucket path with arrow denoting the direction of movement. The blue region denotes the swept area for a normal excavation trajectory.
}
\label{fig: sv}
\end{figure}

Although the amount of soils collected along the trajectory needs to be computed through highly complex simulation of the interaction process between the bucket and soils, we argue that estimating it with the swept volume is admissible provided heavy machinery operations usually do not require high accuracy. Swept volume is defined as the soil volume above the bucket path and calculated through integrating swept areas on the excavation x-z plane along the y axis, as illustrated in Fig.~\ref{fig:sv_illustration}. The swept area $\textsc{A}_{\text{swept}}$ on the x-z plane is numerically computed with trapezoidal rule:
\begin{equation}
    \textsc{A}_{\text{swept}} = \sum_{i=0}^{m-1} \frac{(h_i + h_{i+1}) \Delta x_i}{2}
\end{equation}
where we sample $m$ points on the bucket path; $h_i$ denotes the difference of heights of terrain and the bucket tip at point $i$; $x_i$ is the horizontal distance between point $i$ and point $i+1$.
We then numerically compute the swept volume by sampling $l$ x-z slices from one end point on the bottom plate edge to the other:
\begin{equation}
    \textsc{V}_{\text{swept}} = \sum_{j=0}^{l-1} \textsc{A}_{\text{swept}}^j \Delta y_j
\end{equation}
where $\textsc{V}_{\text{swept}}$ denotes the swept volume, $\textsc{A}_{\text{swept}}^j$ represents swept area on $j$-th slice. Note that terrain surface might have different heights on different slices.

We regulate the trajectory to roughly excavate desired amount of materials through assigning constraints on the normalized bucket fill factor,~\emph{i.e.} the ratio of swept volume to the real bucket volume.
\begin{equation}
    \underline{\textsc{K}}_{bf} \leq \textsc{K}_{bf} = \frac{\textsc{V}_{\text{swept}}}{\textsc{V}_{\text{bucket}}} \leq \overline{\textsc{K}}_{bf}
\end{equation}
where $\textsc{K}_{bf}$ represents the coefficient of bucket fill factor. We adopt an inequality constraint since it is not necessary to stipulate a fixed swept volume. We could control the volume of excavated soils through assigning desired $\textsc{K}_{bf}$ or its range. Ideally, if we want to excavate soils with exact volume of the bucket, we will set $\underline{\textsc{K}}_{bf} = \overline{\textsc{K}}_{bf} = 1$. However, since the estimation is not accurate, we assign $\textsc{K}_{bf}$ according to the observed real collected amount of soils. For example, if we observe the bucket is not full with $\textsc{K}_{bf} = 1$, we will increase the coefficient to achieve the goal of full-bucket filling.

\section{Experiments}
\label{sec: exp}
\begin{figure}
    \centering
    \includegraphics[width=0.85\linewidth]{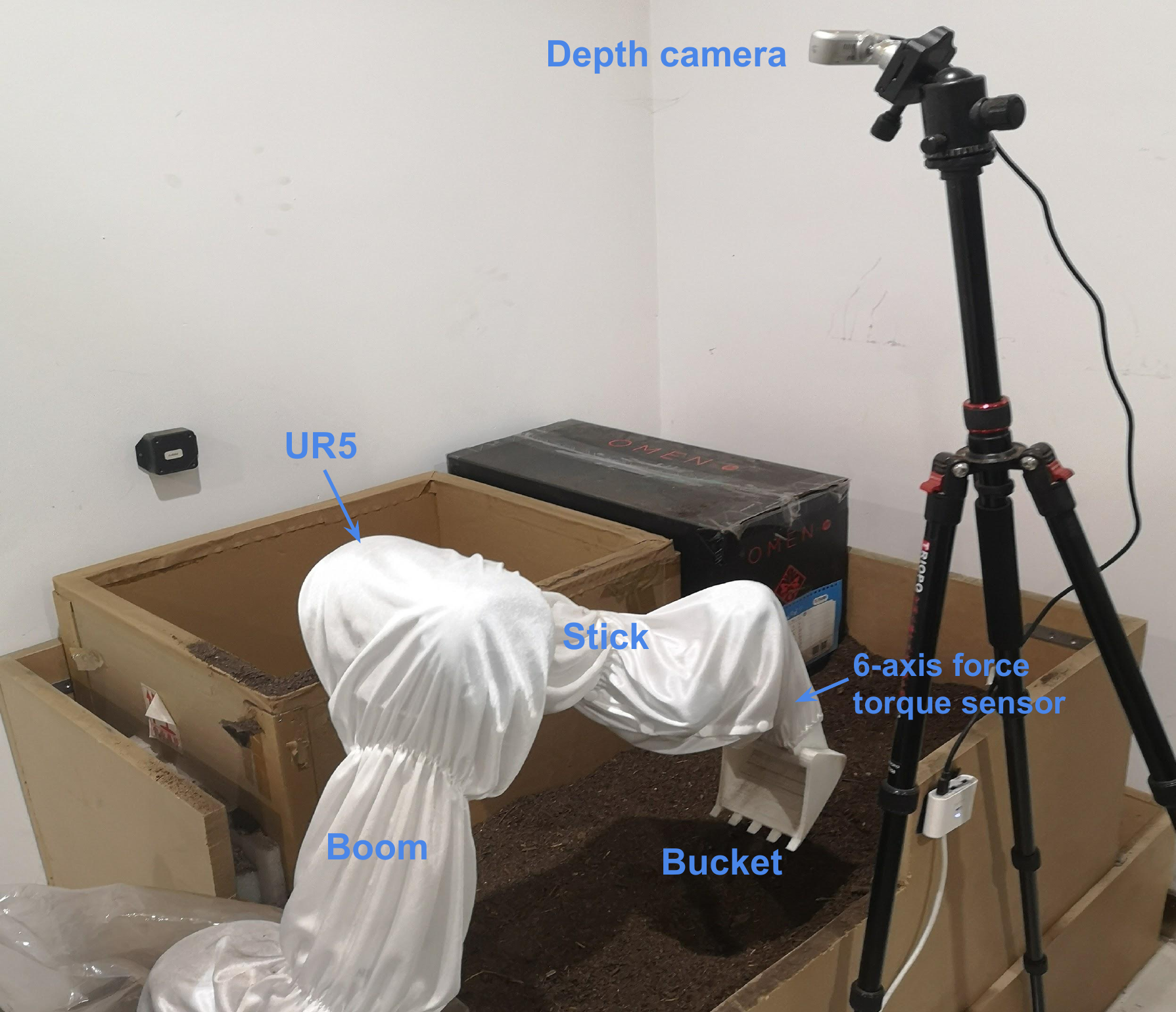}
    \caption{Experiment platform}
    \label{fig: exp_platform}
\end{figure}
We conduct experiments on a real robot platform, where the excavator is substituted by a Universal Robot UR5 with a small bucket, as shown in Fig.~\ref{fig: exp_platform}. Since UR5 has six joints, we fix the last two joints to imitate the $4$ degrees-of-freedom structure of typical excavators. A $6$-axis force-torque sensor connects the end-effector and the bucket, which is used to measure the mass of excavated soils. We use a Intel RealSense depth camera D435 to capture the pointcloud and build the height map of the terrain before every excavation starts. The soil medium is a composite of the pine needle soil that is both low-density and low-cohesive.

\subsection{Swept Volume Estimation Validation}
\label{sec: sv_estimate}
The primary target of an excavation task is to dig sufficient soils out. Our method achieves this target through adding a proper swept volume constraint ($\textsc{K}_{bf}$) that correspond to the desired amount of excavated soils. Therefore, it is crucial to demonstrate the amount of soils actually excavated can be controlled with $\textsc{K}_{bf}$ with sufficient accuracy.
In this experiment, we generate trajectories with different desired $\textsc{K}_{bf}$ to manifest the relation between $\textsc{K}_{bf}$ and the resulting amount of soils actually excavated. We also want to show such relation is valid in a wide variety of excavation conditions, because apart from $\textsc{K}_{bf}$, many other factors could influence the result of an excavation, such as the soil density, the terrain shape and the bucket velocity, etc. For this purpose, we fix the soil density unchanged and allow two different conditions for both terrain shapes and finish time of excavation, which results in four typical conditions:
\begin{enumerate}
    \item flat terrain with finish time of $10$ seconds (Flat-Slow);
    \item slope terrain with finish time of $10$ seconds (Slope-Slow);
    \item flat terrain with finish time of $5$ seconds (Flat-Fast);
    \item slope terrain with finish time of $5$ seconds (Slope-Fast).
\end{enumerate}
For each condition, we conduct a group of experiments with $\textsc{K}_{bf}$ starting from $1$ and increasing until full-bucket filling is achieved. In our cases, full-bucket soils weigh around \SI{1.6}{N}. We repeat $20$ trials for each experiment to learn the stochastic distribution of the mass of excavated soils in terms of $\textsc{K}_{bf}$.

\begin{figure}
\captionsetup[subfigure]{position=b}
\centering
\subcaptionbox{Flat terrain\label{fig: flat_bkt_path}}{\includegraphics[width=.49\linewidth]{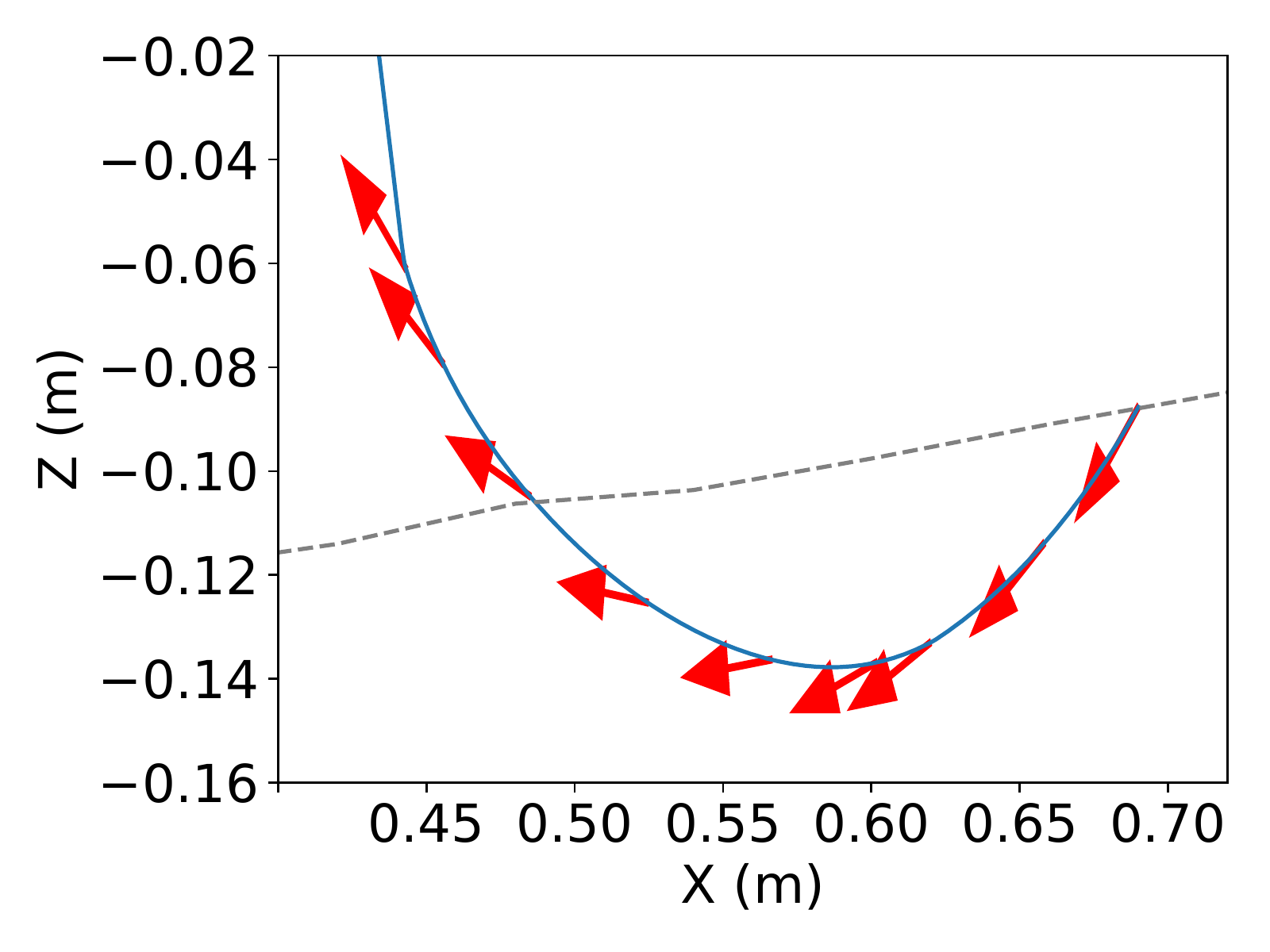}}
\subcaptionbox{Slope terrain\label{fig: slope_bkt_path}}{\includegraphics[width=.49\linewidth]{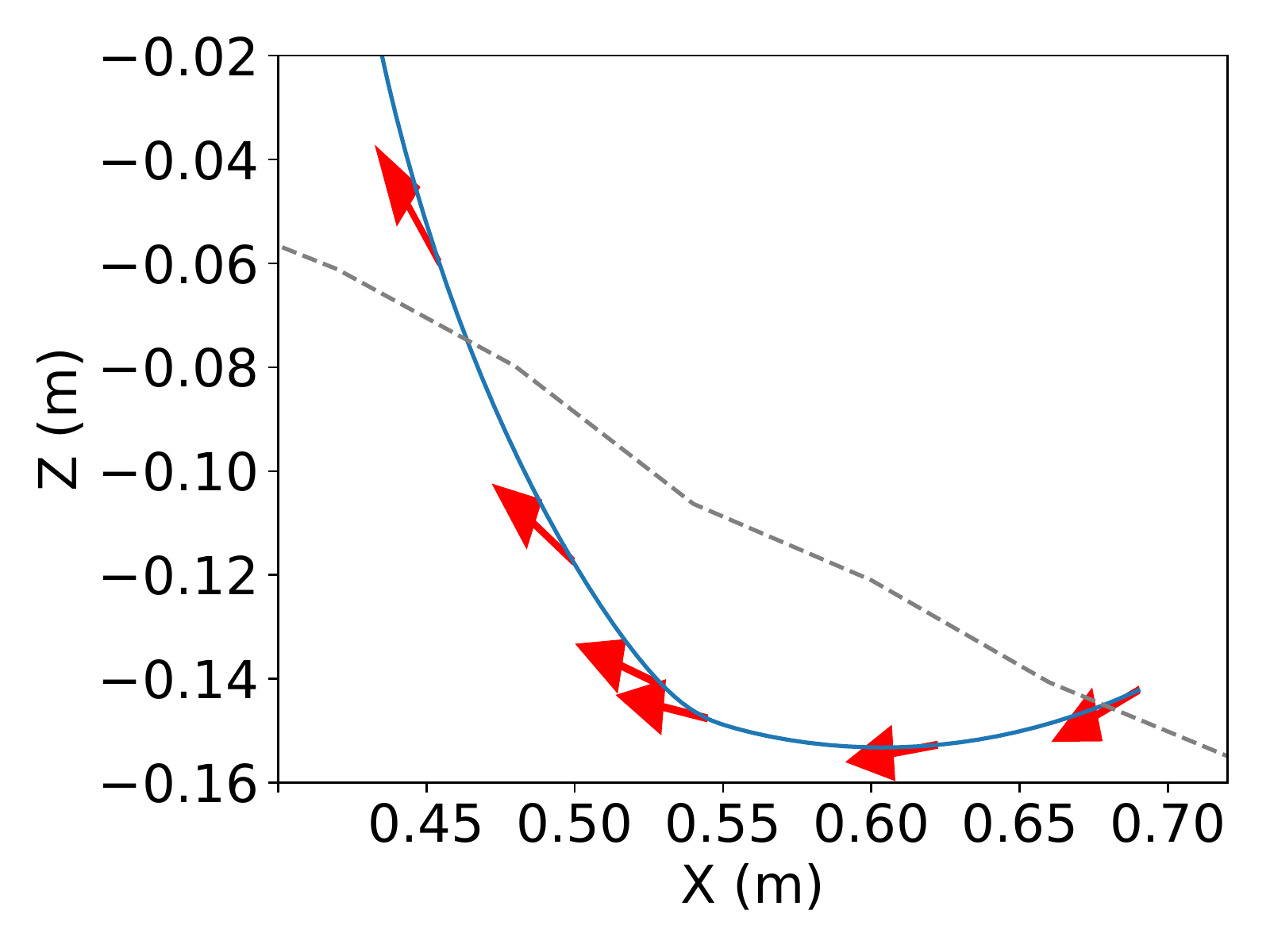}}
\caption{Typical bucket paths generated for flat and slope terrain. Gray dashed lines represent the terrain surfaces. Blue curves are paths of the bucket tip. Red arrows denote heading directions of bucket poses sampled on trajectories.}
\label{fig: bkt_path}
\end{figure}

\begin{figure}
    \centering
    \includegraphics[width=\linewidth]{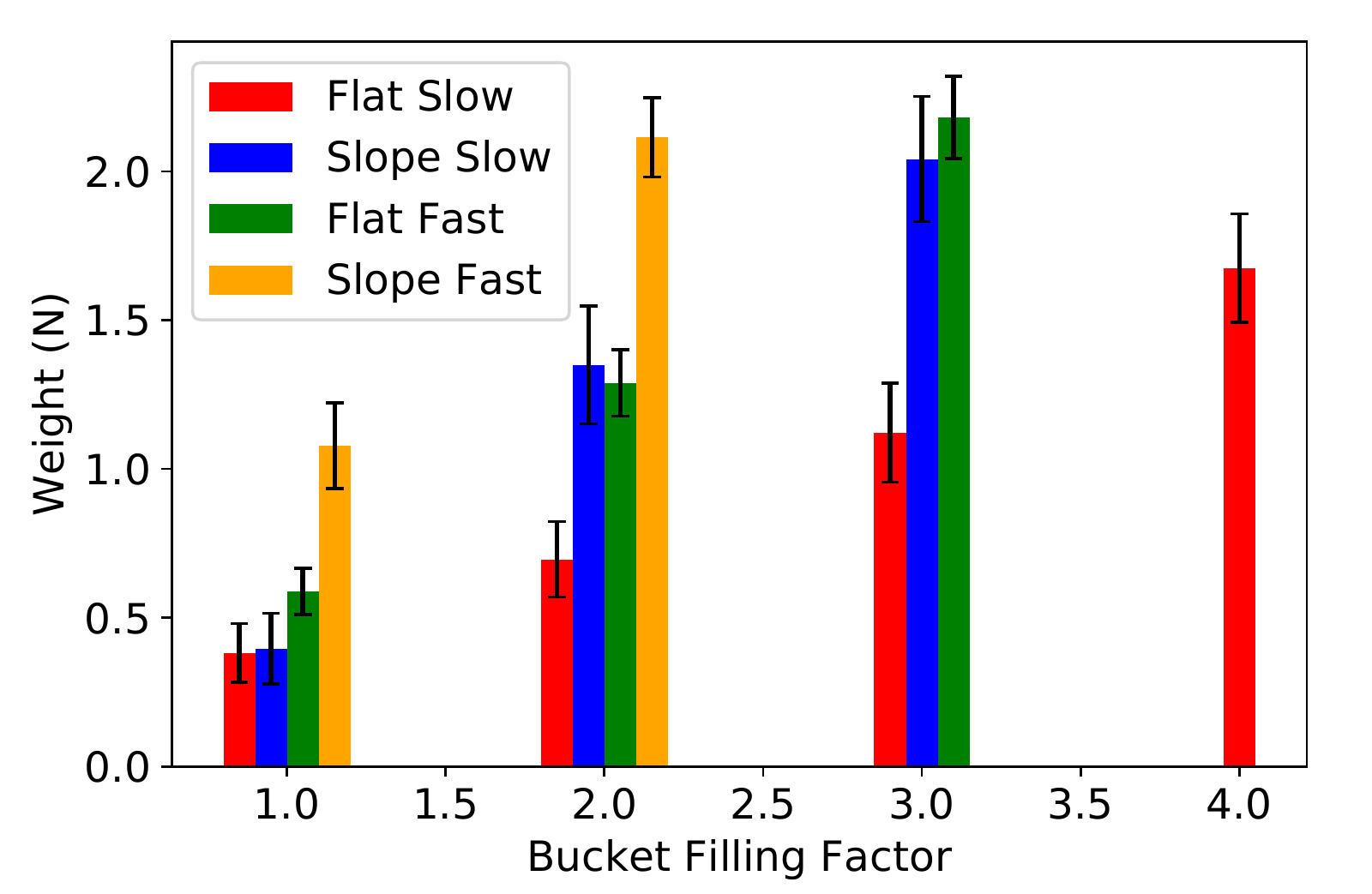}
    \caption{Statistics on weights of excavated soils with different $\textsc{K}_{bf}$. The error bars represent average weights and standard deviations.}
    \label{fig: sv_weights}
\end{figure}

\begin{figure}
\captionsetup[subfigure]{position=b}
\centering
\subcaptionbox{$\textsc{K}_{bf} = 1$\label{fig: sv1}}{\includegraphics[width=.35\linewidth]{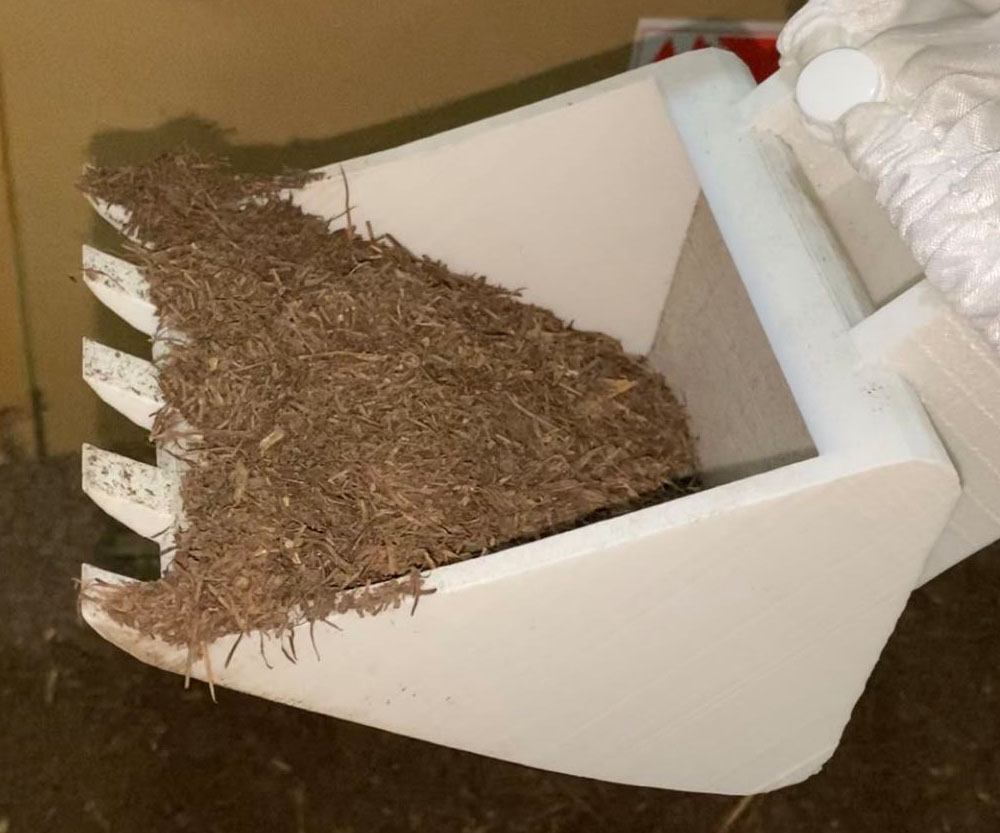}}\hspace{10mm}\vspace*{1mm}
\subcaptionbox{$\textsc{K}_{bf} = 2$\label{fig: sv2}}{\includegraphics[width=.35\linewidth]{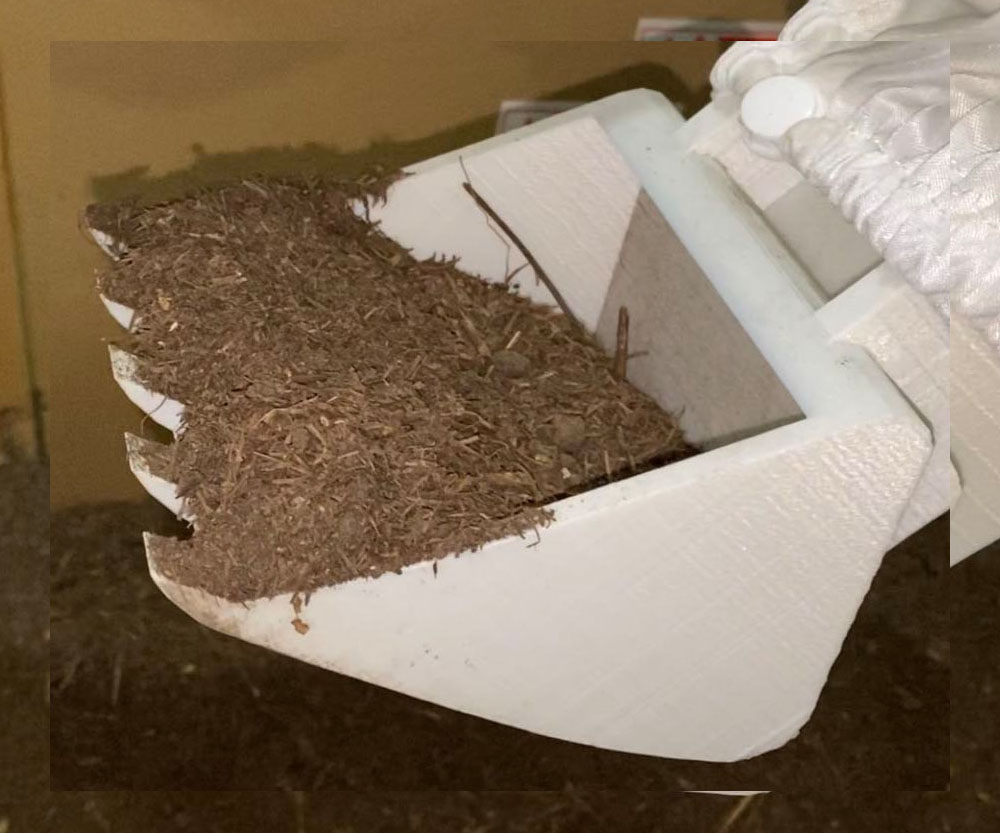}}\vspace*{1mm}
\subcaptionbox{$\textsc{K}_{bf} = 3$\label{fig: sv3}}{\includegraphics[width=.35\linewidth]{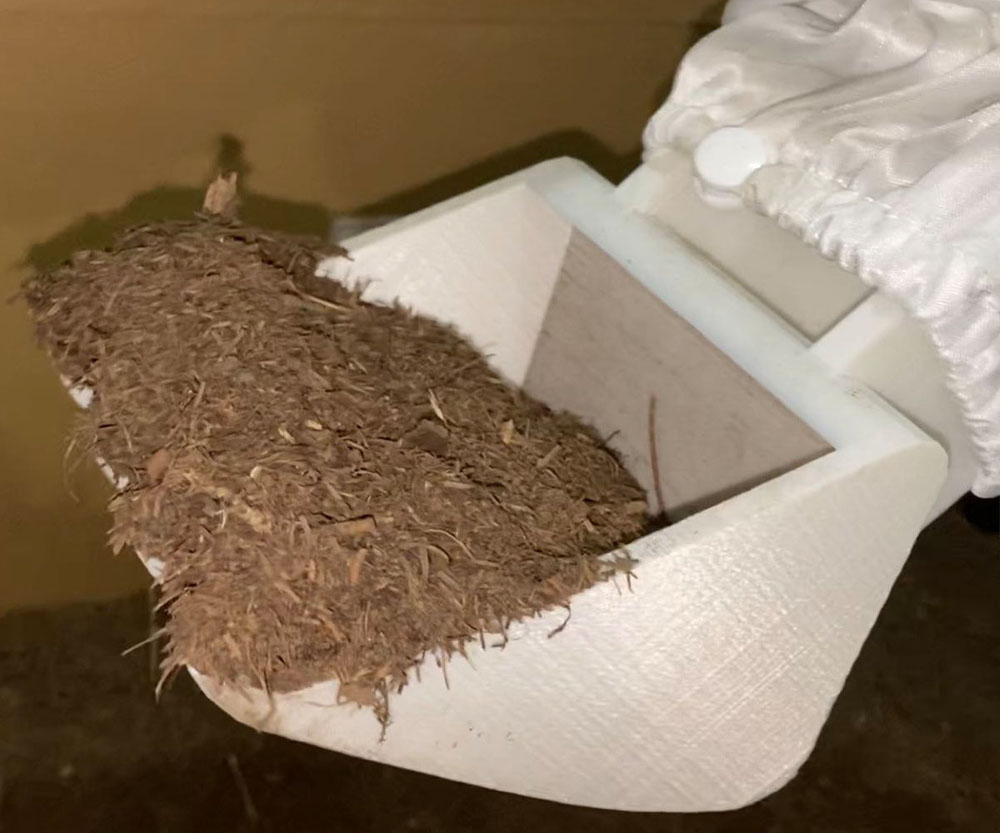}}\hspace{10mm}
\subcaptionbox{$\textsc{K}_{bf} = 4$\label{fig: sv4}}{\includegraphics[width=.35\linewidth]{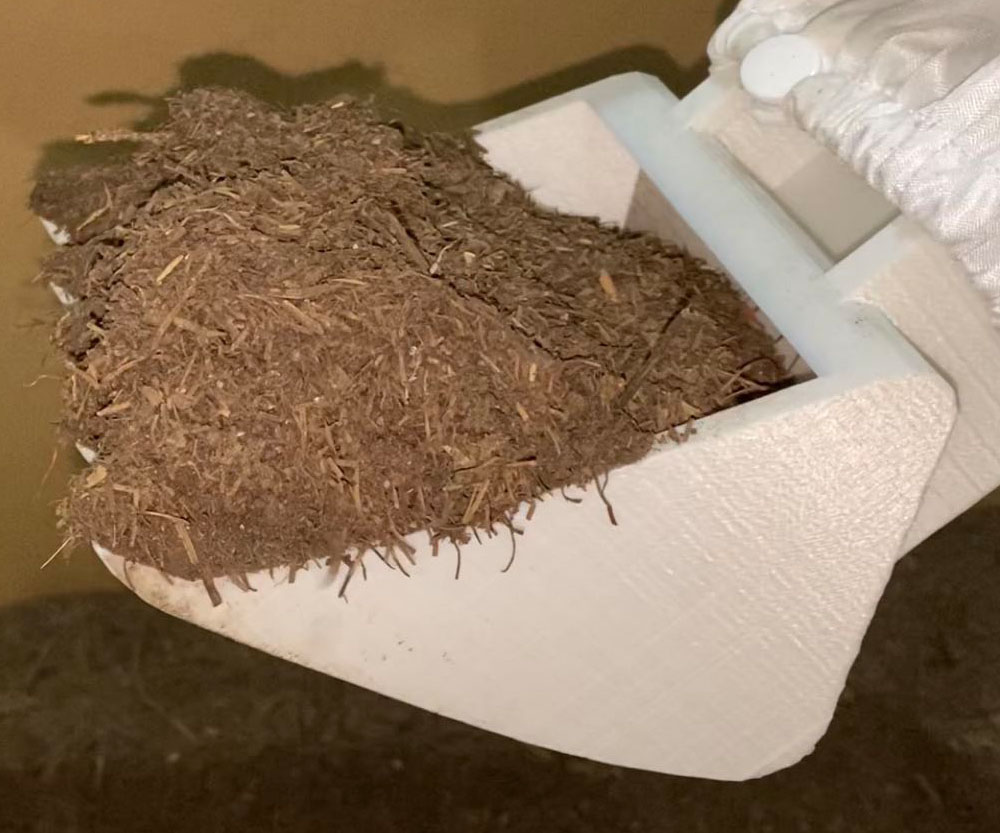}}
\caption{Real amounts of excavated soils with different $\textsc{K}_{bf}$ for Flat-Slow experiments.}
\label{fig: sv_real}
\end{figure}

Fig.~\ref{fig: bkt_path} depicts typical trajectories planned for the flat and slope terrains. Results of the executed excavations are given in Fig.~\ref{fig: sv_weights} and Fig.~\ref{fig: sv_real}. For all conditions, the bucket is not fully filled with $\textsc{K}_{bf} = 1$, which is as expected. This is because some soils are pushed away from the bucket during a practical excavation, while swept volume estimation reduces this complex dynamic process into a static situation. Thus, the swept volume estimation is prone to overestimating the amount of excavated soils. This problem can be solved by increasing $\textsc{K}_{bf}$ based on the fact that, in general, the weight of excavated soils are significantly correlated with $\textsc{K}_{bf}$. However, the exact relation between  $\textsc{K}_{bf}$ and the weight varies for different conditions, because soils experience different dynamic process. As to the trajectory velocity factor, fast trajectories excavate more soils than slow trajectories. One possible reason is that the bucket moves so fast that fewer soils escape from the bucket. As to the terrain shape factor, the bucket could load more soils of the slope terrain than the flat terrain with the same $\textsc{K}_{bf}$. As shown in Fig.~\ref{fig: slope_bkt_path}, for this type of slope, it is hard to push the front soils away since they are pressed by the above soils.



\subsection{Larger Room for Optimization}
\label{sec: room_opt}
\begin{table*}[]
\scriptsize
\centering
\begin{tabular}{|c|c|c|c|c|}
\hline
$\textsc{K}_{bf}$ & pos: $x=\SI{0.6}{m},\ z=\SI{0}{m}$ & pos: $x=\SI{0.8}{m},\ z=\SI{0}{m}$      & pos: $x=\SI{0.6}{m},\ z=\SI{-0.1}{m}$ & pos: $x=\SI{0.8}{m},\ z=\SI{-0.1}{m}$ \\ \hline
1 & 0.0595,\quad \textbf{0.0312} & 0.1287,\quad \textbf{0.0453} & 0.0615,\quad \textbf{0.0332} & 0.1271,\quad \textbf{0.0458} \\ \hline
2 & 0.0764,\quad \textbf{0.0329} & 0.1101,\quad \textbf{0.0433} & 0.0793,\quad \textbf{0.0315} & 0.1100,\quad \textbf{0.0378} \\ \hline
3 & 0.0835,\quad \textbf{0.0339} & 0.1388,\quad \textbf{0.0356} & 0.0909,\quad \textbf{0.0339} & 0.1288,\quad \textbf{0.0365} \\ \hline
\end{tabular}
\caption{Comparison of minimal length of joint trajectory using the three-phase method and our method. Results of our method are highlighted in bold type.}
\label{tab: comp_min_length}
\end{table*}
This experiment is to demonstrate that our constraint-based task specification allows larger room for optimization, by comparing the minimal length of joint trajectory obtained by our method and that obtained by exhaustive searching parameters of the classical three-phase specification. For a typical three-phase excavation~\cite{sing1995synthesis}, its parameterization for linear penetration, dragging and curling is shown in~\cite{sing1995synthesis} with the penetration length $l_p$, penetration angle $\alpha_p$, dragging length $l_d$, and curling radius $r$. Since the start location of an excavation with respect to the excavator's base greatly influences joint movements, we conduct four groups of experiments with different starting positions. For each group of experiments, we further conduct three trials with different desired $\textsc{K}_{bf}$. Table~\ref{tab: comp_min_length} shows experiment configurations and results, where $(x, z)$ coordinates of starting positions are given in the first row and values of desired $\textsc{K}_{bf}$ are given in the first column, and minimal lengths of joint trajectories computed according to Eq.~\ref{eq: jnt_pos_obj} are highlighted in bold.
\begin{figure}
    \centering
    \includegraphics[width=.5\linewidth]{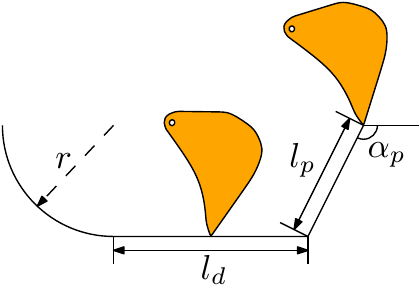}
    \caption{Parameterization of three-phase specification.}
    \label{fig: three_phase_param}
\end{figure}

\begin{figure}
\captionsetup[subfigure]{position=b}
\centering
\subcaptionbox{Joint Trajectory\label{fig: min_length_jnt}}{\includegraphics[width=.8\linewidth]{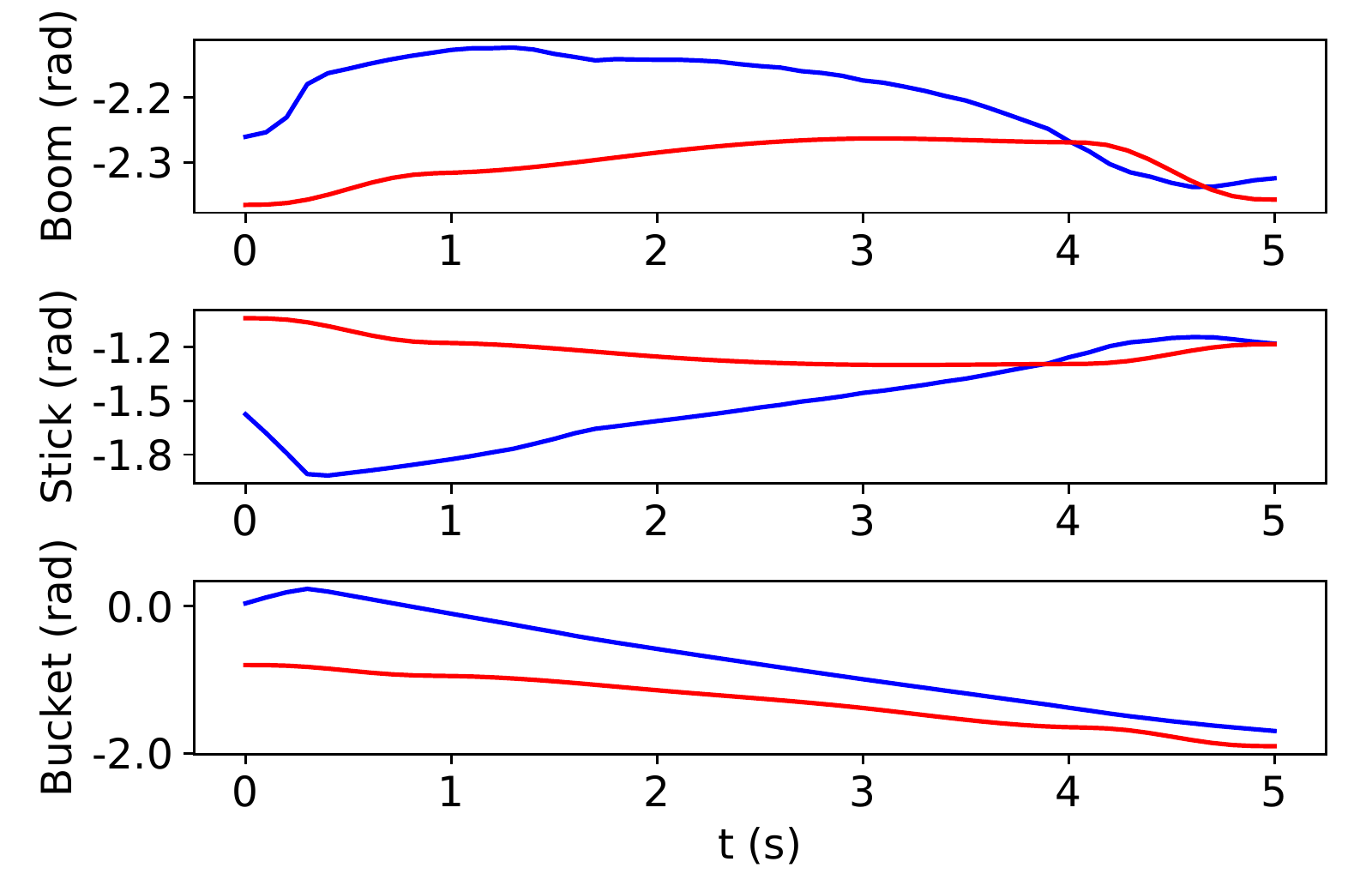}}
\subcaptionbox{Bucket Path\label{fig: min_length_bkt}}{\includegraphics[width=.8\linewidth]{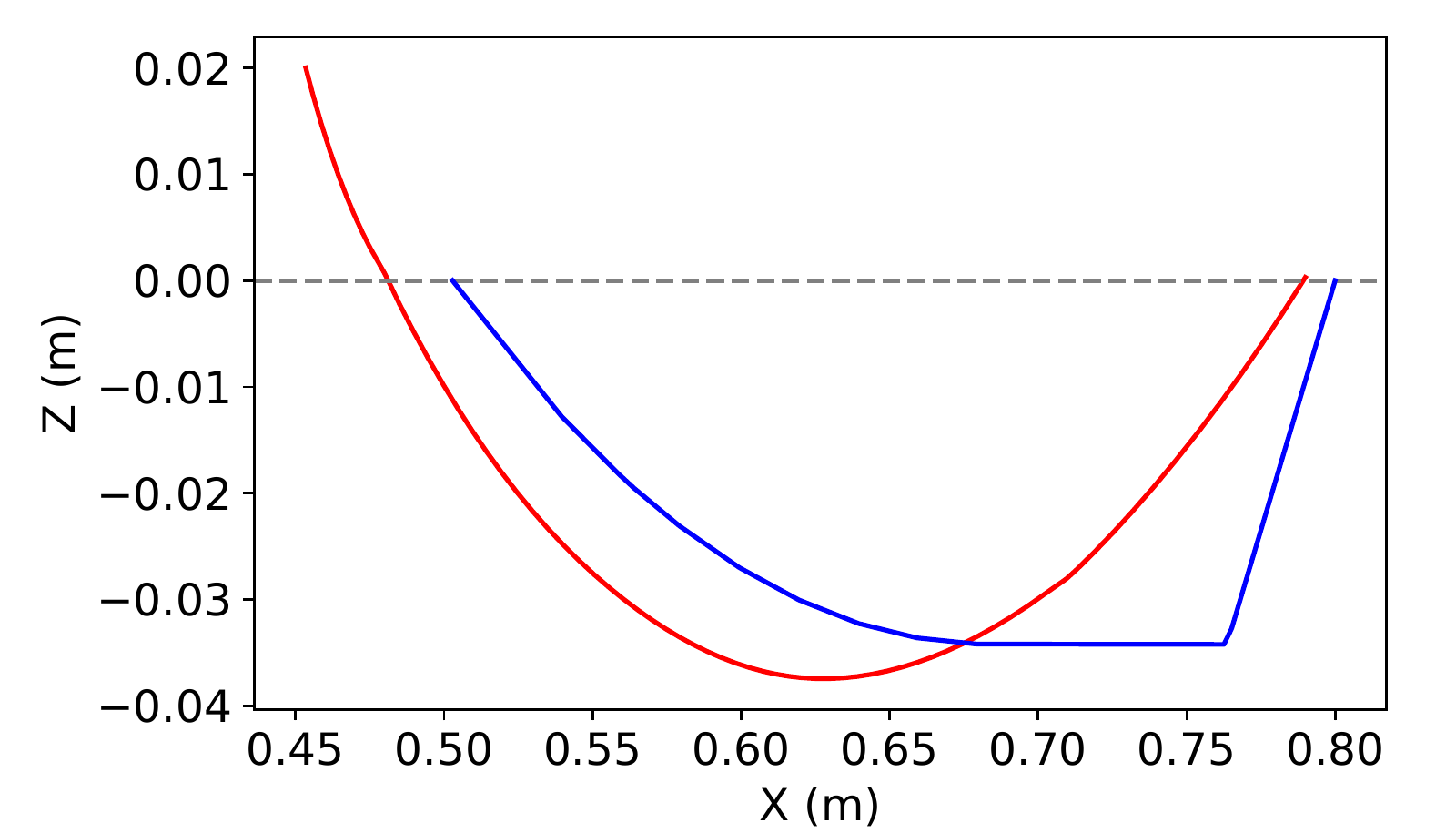}}
\caption{Minimal length trajectories for starting position $x = 0.8, z = 0$ and $\textsc{K}_{bf} = 3$. Red trajectories are generated by our method; Blue trajectories are generated by the three-phase method.}
\label{fig: min_length_traj}
\end{figure}

Results show that, our method is able to find a shorter trajectory for all conditions. The fundamental reason is that the linear translation in three-phase actions requires simultaneous movements of multiple joints in the excavator, which is determined by the mechanical structure of the excavator. However, large multi-joint movements are usually unnecessary to dig soils out. This can be verified by Fig.~\ref{fig: min_length_jnt}: from $1$s to $4$s, the trajectory of the three-phase method experiences a large three-joint movement, while for our method the boom and stick only move subtly. And comparison of the corresponding bucket paths is depicted in Fig.~\ref{fig: min_length_bkt} and it is obvious that our method can provide a smoother bucket trajectory.

\begin{table*}[]
\centering
\resizebox{\textwidth}{!}{%
\begin{tabular}{|c|c|c|c|c|c|c|c|c|c|c|c|c|}
\hline
\multirow{2}{*}{$\textsc{K}_{bf}$} & \multicolumn{3}{c|}{ pos: $x=\SI{0.6}{m},\ z=\SI{0}{m}$} & \multicolumn{3}{c|}{pos: $x=\SI{0.8}{m},\ z=\SI{0}{m}$} & \multicolumn{3}{c|}{pos: $x=\SI{0.6}{m},\ z=\SI{-0.1}{m}$} & \multicolumn{3}{c|}{pos: $x=\SI{0.8}{m},\ z=\SI{-0.1}{m}$} \\ \cline{2-13} 
 & Min & Mean & Std Dev & Min & Mean & Std Dev & Min & Mean & Std Dev & Min & Mean & Std Dev \\ \hline
1 & 6.19, \textbf{4.50} & 6.83, \textbf{5,78} & 0.48, \textbf{0.55} & 6.07, \textbf{5.04} & 8.80, \textbf{6.34} & 2.17, \textbf{0.75} & 5.68, \textbf{5.11} & 7.46, \textbf{6.25} & 1.04, \textbf{0.52} & 5.89, \textbf{4.69} & 8.25, \textbf{6.19} & 1.39, \textbf{0.91} \\ \hline
2 & 6.08, \textbf{5.35} & 7.53, \textbf{6.44} & 1.51, \textbf{0.77} & 6.22, \textbf{5.84} & 8.71, \textbf{6.79} & 1.60, \textbf{0.54} & 5.97, \textbf{4.89} & 7.64, \textbf{6.27} & 1.19, \textbf{0.91} & 6.00, \textbf{5.58} & 8.22, \textbf{6.81} & 1.49, \textbf{0.67} \\ \hline
3 & 6.05, \textbf{5.58} & 7.41, \textbf{6.91} & 0.94, \textbf{0.52} & 6.91, \textbf{5.31} & 8.93, \textbf{6.83} & 1.53, \textbf{0.76} & 6.43, \textbf{5.18} & 7.82, \textbf{6.72} & 1.10, \textbf{0.69} & 5.92, \textbf{4.51} & 8.47, \textbf{6.64} & 1.53, \textbf{0.83} \\ \hline
\end{tabular}%
}
\caption{Comparison of results of time optimal trajectory generation using one-stage method and two-stage method. Min stands for the minimal time out of all generated trajectories; Mean denotes the average time; Std Dev is the standard deviation. Results of our method are highlighted in bold type.}
\label{tab: comp_min_time}
\end{table*}
\subsection{Time Optimal Trajectory Generation}
We compare results of our one-stage method and the traditional two-stage method. The two-stage algorithm is implemented as follows:
\begin{enumerate}
    \item We set the minimal travel time as the objective in the optimization formulation;
    \item We first compute a path satisfying task constraints without objective function. Then we scale time intervals for this path using time optimal trajectory generation technique proposed in~\cite{kunz2012time}.
\end{enumerate}
As in~\ref{sec: room_opt}, we conduct $12$ groups of experiments with $4$ different starting positions and $3$ different $\textsc{K}_{bf}$. For all conditions, we try multiple initial trajectories with all initial time intervals of \SI{2}{s} to search the minimal time trajectory among them. 

Results are shown in Table~\ref{tab: comp_min_time}. For all conditions with multiple different initial trajectories, the minimal time of the one-stage method is shorter than that of the two-stage method by $16\%$. The improvement of the one-stage method in average time is more significant, which is $18\%$ shorter than that of the two-stage method. Besides, the standard deviation of the one-stage timing cost is smaller than that of the two-stage method by $42\%$.

\subsection{Productivity Test}
\begin{figure*}
    \centering
    \includegraphics[width=\linewidth]{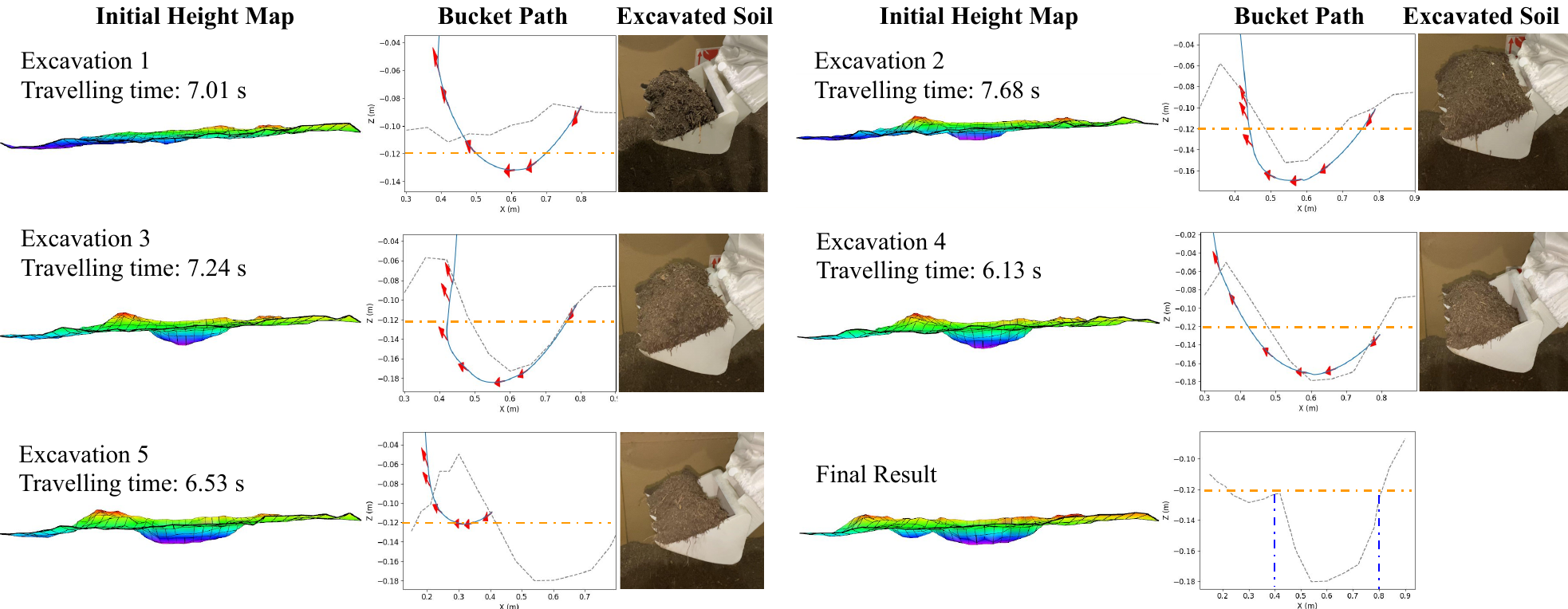}
    \caption{Five continuous excavations in the productivity test. Blue dashed lines show the region that we consider. Orange dashed lines are the target height. Best viewed on screen.}
    \label{fig: prod_test}
\end{figure*}

This experiment is to demonstrate that our method can generate trajectories with high productivity and adaptability to terrains in different shapes. The task requires the excavator to efficiently remove soils in a region until the surface of this local terrain is lower than a target height. Because of the limited space of robot motion, we consider a line region that $x \in [\SI{0.4}{m}, \SI{0.8}{m}]$ and $y = \SI{0}{m}$. The initial height of the local terrain is around $\SI{-0.1}{m}$, and the target height is $\SI{-0.12}{m}$. For each excavation, we generate a time-optimal trajectory with the joint velocity constraint that every joint can not exceed the speed of $\SI{0.3}{rad/s}$. As the excavation progresses, the shape of the terrain largely changes. To guarantee that sufficient soils are dug out in every iteration no matter what the terrain shape is, we conservatively assign $\textsc{K}_{bf} = 3$ according to the experiment results in Section~\ref{sec: sv_estimate}. Fig.~\ref{fig: prod_test} illustrates the five continuous excavations to fulfill the task. 

\section{Conclusion}
We develop an optimization-based framework of excavation trajectory generation, which allows optimizing various criteria,~\emph{e.g.} the length of trajectory and the total travelling time, while guaranteeing digging sufficient soils out for terrains with different shapes. In this framework, we propose a set of flexible constraints to specify the excavation task by regulating instantaneous motions of the bucket. As a result, the optimization can be performed over a broader trajectory space compared to the previous work and can converge to better solutions. Besides, this framework facilitates time-optimal trajectory generation by introducing time variables and optimizing the total travelling time in one stage.

Experiments conducted on a real robot platform confirms that trajectories generated by our method are practical to excavate sufficient soils and adaptive to different terrains. The advantage of larger room of optimization is demonstrated by the experiment result that the minimal-length trajectory generated by our method is apparently shorter than that exhaustively searched in a simplified action space. As to the time-optimal trajectory generation, our experiments show that one-stage time-variable trajectory optimization is worth exploring because of the better performance on execution time. However, we must admit that introduction of time variables makes the optimization problem more difficult in terms of more optimization variables and nonlinearity. Thus, the rate of successful optimization of the one-stage method is less than that of the two-stage method.

There remains a lot of future work related to the optimal excavation trajectory generation. Our framework currently estimates the amount of excavated soils with static swept volume, but it can easily extend to other accurate models of interaction between soils and the bucket, which is possibly the most challenging task in the excavation field. An alternative approach is to develop an adaptive strategy to adjust the desired bucket filling factor according to properties such as the shape of the terrain and attributes of the soil.

\bibliographystyle{IEEEtran}
\bibliography{reference}

\end{document}